\documentclass[conference]{IEEEtran}

\IEEEoverridecommandlockouts
\usepackage[utf8]{inputenc}
\usepackage[T1]{fontenc}
\usepackage{latexsym}
\usepackage{url}
\usepackage[numbers,sort&compress]{natbib}
\usepackage{booktabs}
\usepackage{amsmath,amssymb,amsfonts}
\usepackage{graphicx}
\usepackage{xcolor}
\usepackage{multirow}
\usepackage{array}
\usepackage{float}
\usepackage{placeins}
\usepackage{microtype}
\usepackage{tikz}
\usetikzlibrary{arrows.meta,positioning}

\newcommand{\method}{\textsc{RSP}}
\newcommand{\dataset}[1]{\textsc{#1}}
\newcommand{\randomfour}{Random@\(4B\)}
\newcommand{\droponly}{\textsc{RSP-Drop}}

\title{Perturbation-Based Data Selection for Medical Question Answering}

\author{%
\IEEEauthorblockN{Yuexin Wu \quad Dayou Yu \quad Vasile Rus}
\IEEEauthorblockA{Department of Computer Science\\
University of Memphis, Memphis, TN, USA\\
\texttt{ywu10@memphis.edu \quad dayou.yu@memphis.edu \quad vrus@memphis.edu}}
\thanks{This work has been submitted to IEEE for possible publication. Copyright may be transferred without notice, after which this version may no longer be accessible.}}


\begin{document}
\maketitle
\raggedbottom

\begin{abstract}
Selecting a compact training subset is an important step in adapting medical language models from existing question-answering banks. We investigate perturbation-based data selection using two signals derived from Gaussian perturbations of the final-block representation: prediction-change frequency and non-flipping margin contraction. After a shared entropy prefilter, Representation-Sensitivity Prioritization (RSP) combines the signals using automatic root-mean-square scaling, while RSP-Drop ranks candidates by contraction alone. Both selectors use only questions and answer options during selection. We evaluate them with MedGemma-4B on five medical QA datasets at retained-set budgets of 64, 128, and 256 examples. At the primary budget of 128, RSP and RSP-Drop obtain dataset-averaged Macro-F1 scores of 47.55 and 48.28, respectively, compared with 45.11 for entropy-pool-matched random selection. Their relative performance varies across datasets and budgets. A selected-set case study identifies contrasting clean-margin profiles among the examples exchanged by the two rules, while a full-pool ablation shows that entropy prefiltering is not uniformly beneficial. These retrospective results characterize the empirical utility and limitations of perturbation-based selection under a fixed adaptation protocol. Evaluation uses one backbone and benchmarks with documented training-family overlap; it does not establish clinical effectiveness.

\end{abstract}

\section{Introduction}
\label{sec:introduction}

Developers adapting a medical foundation model may already possess an authorized bank of medical questions, answer labels, and auxiliary explanatory texts, while a planned update can retain only a fixed number \(B\) of training examples. This creates an offline data-selection problem: which examples should enter the final update under the same retained-set budget? The selector operates on question-and-option model behavior before auxiliary text is introduced; the downstream adaptation recipe is held fixed when comparing selection rules. The retained-set budget is distinct from the cost of creating, validating, scoring, or tuning the candidate bank.

Common selection heuristics emphasize output uncertainty or representation coverage. Entropy and margin identify uncertain predictions; BALD-style criteria use stochastic disagreement \citep{houlsby2011bayesian,kirsch2019batchbald}; and diversity methods seek broader feature-space coverage \citep{hubotter2025sift}. These signals do not directly ask how a local change to a question representation affects the answer boundary. We examine two selectors built from these responses: \method{} combines prediction changes with non-flipping margin contraction, whereas \droponly{} uses contraction alone after the same entropy prefilter. Their comparison asks whether adding Flip-prioritized examples improves adaptation relative to continuous-drop ranking.

Our goal is to characterize when these two operational sensitivity rules help or fail under a fixed protocol, not to establish a universally optimal selector or retrospectively choose a dataset-specific winner. Both selectors are auxiliary-text-blind and label-blind: the raw model receives only a question and its options. For \method{}, the mixing coefficient is computed from unlabeled score components in the prefiltered pool. Labels and auxiliary text are introduced only after top-\(B\) membership is frozen. We make three contributions:

\begin{itemize}
    \item We define two auxiliary-text-blind selectors: \method{} with automatic Flip--Drop scale balancing, and entropy-prefiltered \droponly{} with direct contraction ranking.
    \item We compare both selectors with uncertainty, diversity, and auxiliary-text-aware comparators on five medical QA datasets at \(B=128\), with budget analyses at \(B=64,128,256\).
    \item We combine selected-set and score-support analyses with an entropy-prefilter ablation to characterize how component utility varies across datasets, budgets, and candidate-pool restrictions.

\end{itemize}

Code, experiment configurations, and aggregate results for this extended study are publicly available.\footnote{\url{https://github.com/woqingdoua/rsp-medical-qa-selection-ichi}}

\paragraph{Relationship to the Prior Workshop Study}
This study extends the non-archival GenAI4Health @ NeurIPS 2026 workshop study of development-tuned representation sensitivity~\citep{wu2026development}. That study combined Flip frequency and non-flipping margin contraction with a coefficient selected using labeled development performance. We retain these underlying signals and the entropy-prefiltered selection framework, replace the development-based coefficient search with automatic RMS scale balancing, and introduce \droponly{} as a contraction-only variant. We additionally examine retained-set budgets of \(B=64,128,256\), score support, selected-set differences, and entropy-prefilter ablations. Both studies use the same five benchmark families and MedGemma backbone; these extensions therefore broaden the analysis within that setting rather than establish generalization to new benchmarks or backbones. Cross-paper numerical differences are not a controlled estimate of the benefit of automatic scaling; the within-study comparisons reported here provide the basis for our empirical conclusions.

\paragraph{Intended use}
We evaluate offline curation of an existing medical-QA bank under a fixed retained-set budget and adaptation recipe. The contribution is an empirical comparison across benchmarks, budgets, and components, not patient-level decision making or clinical deployment. Label-blind scoring still requires labeled training data; neither question/options-only selector assesses explanation quality.

\section{Related Work}
\label{sec:related}

\paragraph{Auxiliary Explanations During Adaptation}
Chain-of-thought (CoT) prompting can expose useful intermediate reasoning on multi-step tasks \citep{wei2022chain}, but a generated or collected explanation need not faithfully reflect the process that produced an answer \citep{turpin2023unfaithful,lanham2023faithfulness}. Our datasets contain heterogeneous auxiliary text---including reference explanations and an abstract conclusion---so we avoid treating every field as human CoT. We select examples from a pre-existing bank and supervise only the final answer.

\paragraph{Data selection}
Entropy, margin, and BALD quantify different forms of uncertainty \citep{houlsby2011bayesian}; batch-disagreement and representation-diversity criteria provide alternative selection rules \citep{kirsch2019batchbald,hubotter2025sift}. Neuron-activation signals provide a complementary representation-based signal \citep{chen2026nait}. Our comparison also includes explanation-aware RSR-, ASLEC-DROP-, and TEMP-inspired selectors \citep{yang2026rsr,wang2026aslec,jin2026temp}. These differ from question/options-only scoring in their access to auxiliary text. We explicitly identify our implementations as protocol-specific adaptations to an existing labeled bank and a common LoRA endpoint, rather than exact reproductions of the cited methods.

\paragraph{Perturbation signals}
Prediction stability under noise complements uncertainty and diversity. Our Noise-Stability comparator constructs a hidden-sensitivity representation before diversity selection. \method{} instead preserves the direction of the answer-boundary response by separating Flip from non-flipping margin contraction. We treat both as operational heuristics, not calibrated measures of uncertainty or perturbation severity.

\section{RSP and RSP-Drop}
\label{sec:method}

\subsection{Problem Setup}

Let \(\mathcal{C}=\{(x_i,y_i,r_i)\}_{i=1}^{N}\) be a bank containing a question and its options \(x_i\), an answer-position label \(y_i\), and auxiliary explanatory text \(r_i\). A selector returns \(\mathcal{S}\subset\mathcal{C}\), \(|\mathcal{S}|=B\). During scoring for either selector, the model receives only \(x_i\): neither \(y_i\) nor \(r_i\) is read. The UID is used only for bookkeeping and selection-signature checks; it is not a semantic model input. After selection, every retained training example contains \(r_i\) in its context and places loss only on final-answer target tokens. Evaluation again presents only \(x_i\), without \(r_i\).

The \method{} coefficient \(\lambda_0\) uses only unlabeled candidate scores, without adapter fitting or development evaluation. The retained-item budget does not price explanation creation, scoring, or example length.

\subsection{Shared Perturbation Signals}

Scoring starts from the unadapted \texttt{google/medgemma-4b-it} checkpoint; no warm-up adapter is loaded. For item \(i\), let \(H_i^{(L)}\in\mathbb{R}^{T_i\times d}\) be the complete output tensor of the final decoder block, before subsequent model-level normalization and the language-model head. Each stochastic pass adds independent absolute Gaussian noise to every sequence position and hidden dimension:
\begin{equation}
\widetilde{H}_i^{(L,k)}=H_i^{(L)}+\epsilon_i^{(k)},
\qquad
\epsilon_i^{(k)}\sim\mathcal{N}(0,\sigma^2 I).
\label{eq:perturb}
\end{equation}
Although noise is instantiated over the stored final-block output tensor, subsequent normalization and the language-model head are token-wise, and the score uses next-token logits at the final prompt position. Consequently, only perturbation at that position affects the restricted answer probabilities; nonfinal-position noise cannot propagate through later attention because the intervention follows the final decoder block. Operationally, this is a final-position output-representation perturbation. Let \(z_{i,T_i}\) be the final-position vocabulary logits, \(\mathcal{A}_i\) the legal option letters, and \(v_y\) the single tokenizer ID for the string \texttt{" y"}. The clean distribution is
\begin{equation}
p_i(y)=\frac{\exp z_{i,T_i}(v_y)}
{\sum_{c\in\mathcal{A}_i}\exp z_{i,T_i}(v_c)},
\qquad y\in\mathcal{A}_i,
\label{eq:option-prob}
\end{equation}
and \(\widetilde p_i^{(k)}\) is defined analogously after Eq.~\eqref{eq:perturb}. This is a restricted next-token softmax over legal option letters, not a sequence probability for the option text. The implementation requires each leading-space letter to be one token; multi-token answer labels are unsupported.

For each candidate, we compare the clean prediction with predictions obtained after adding Gaussian perturbations to its final-block representation. A Flip occurs when the perturbed prediction differs from the clean prediction. Let \(\widehat y_i=\arg\max_y p_i(y)\). Exact probability ties resolve to the earliest legal letter (A before B, and so on). We define the Flip frequency
\begin{equation}
\operatorname{Flip}(x_i)=\frac{1}{K}\sum_{k=1}^{K}
\mathbf{1}\!\left[\arg\max_y\widetilde p_i^{(k)}(y)\ne\widehat y_i\right].
\label{eq:flip}
\end{equation}
If \(p_{i,(1)}\) and \(p_{i,(2)}\) are the two largest clean probabilities, then \(m_i^{(0)}=p_{i,(1)}-p_{i,(2)}\). For pass \(k\), the clean winner's margin is
\begin{equation}
m_i^{(k)}=\widetilde p_i^{(k)}(\widehat y_i)
-\max_{y\ne\widehat y_i}\widetilde p_i^{(k)}(y),
\end{equation}
and the non-flipping contraction is
\begin{equation}
\begin{split}
\operatorname{Drop}(x_i)=\frac{1}{K}\sum_{k=1}^{K}
&\mathbf{1}\!\left[\arg\max_y\widetilde p_i^{(k)}(y)=\widehat y_i\right]\\
&{}\cdot\max\!\left(0,m_i^{(0)}-m_i^{(k)}\right).
\end{split}
\label{eq:drop}
\end{equation}
\subsection{RSP: Automatic RMS Scale Balancing}

The two signals have different empirical scales: \(\operatorname{Flip}\) is a frequency in increments of \(1/K\), while \(\operatorname{Drop}\) is a probability-margin contraction. We first retain the \(\min(4B,N)\) examples with highest predictive entropy, using a stable sort, and denote this candidate pool by \(\mathcal{P}\), with \(M=|\mathcal{P}|\). Abbreviate
\(F_i=\operatorname{Flip}(x_i)\) and
\(D_i=\operatorname{Drop}(x_i)\). We compute the root-mean-square scale of each component within \(\mathcal{P}\):
\begin{align}
a_F &= \left(\frac{1}{M}\sum_{i\in\mathcal{P}}F_i^2\right)^{1/2},
&
a_D &= \left(\frac{1}{M}\sum_{i\in\mathcal{P}}D_i^2\right)^{1/2},
\label{eq:rms-scales}
\end{align}
The automatic coefficient and final priority score are
\begin{equation}
\lambda_0=\frac{a_F}{a_D+\varepsilon},
\qquad \varepsilon=10^{-8},
\label{eq:auto-lambda}
\end{equation}
\begin{equation}
s_{\mathrm{RSP}}(x_i)=
\begin{cases}
F_i+\lambda_0D_i, & \lambda_0>0,\\
D_i, & \lambda_0=0.
\end{cases}
\label{eq:rsp}
\end{equation}
When \(a_D\gg\varepsilon\), the RMS magnitude of \(\lambda_0D_i\) approximately matches that of \(F_i\). This corrects the component-scale mismatch without a coefficient search, but does not guarantee equal influence on the ranking. The coefficient is computed separately for each dataset--budget--acquisition-seed pool after entropy prefiltering and is frozen before top-\(B\) selection. Computing \(\lambda_0\) reads neither answer labels nor auxiliary text and requires no adapter fitting or development evaluation.

For a zero coefficient, candidates are ranked directly by the Drop component; exact score ties use the same original-pool-index rule. The implementation tests \(a_F\le10^{-8}\). With \(K=8\), \(M\le1024\), and positive \(F_i\ge1/8\), any positive \(a_F\) is at least \(1/(8\sqrt M)\ge1/256\); this threshold therefore identifies the zero-coefficient case in the reported entropy-prefiltered settings. All 45 dataset--budget--seed records have positive coefficients and no all-candidate score collapse; unique priority counts range from 89 to 402. The retained historical implementation evaluates the positive-coefficient expression directly, which gives the same scores on these runs. Partial cutoff ties are reported in RQ3.

\subsection{RSP-Drop: Direct Contraction Ranking}
\label{sec:drop-only}
The contraction-only variant uses the identical entropy-restricted pool \(\mathcal P\), clean predictions, perturbation layer, noise scale, and \(K\) passes, but ranks candidates by
\begin{equation}
s_{\mathrm{Drop}}(x_i)=D_i.
\label{eq:drop-selector}
\end{equation}
It retains the top \(B\) candidates, resolving exact ties by original pool index. We call this variant \droponly{} throughout. It is not a full-pool selector unless \(\min(4B,N)=N\). It requires neither a mixing coefficient nor an additional scoring model. Since Eq.~\eqref{eq:drop} divides by all \(K\) passes, rather than only non-flipping passes, a candidate that flips on every pass has \(D_i=0\).
\method{} and \droponly{} express different selection preferences: the former includes an explicit reward for prediction changes, while the latter ranks only non-flipping contractions. Neither directly estimates the downstream benefit or quality of an explanation. Both are reported as fixed methods; we do not select between them using test performance.

\subsection{Frozen Acquisition Protocol}

We fix \(K=8\), the final decoder block, \(\sigma=0.02\), and entropy-prefilter multiplier four. The primary budget is \(B=128\), with \(B=64,256\) in the retrospective budget evaluation. For each dataset--budget--acquisition-seed pool, we compute Eqs.~\eqref{eq:rms-scales}--\eqref{eq:rsp}, rank by decreasing \(s_{\mathrm{RSP}}\) or \(s_{\mathrm{Drop}}\), and retain \(B\) examples. Exact ties resolve by original pool-row index; training restores ascending pool-index order, so ranking changes membership, not presentation order. These settings are fixed design choices: there is no \(\lambda\)-grid, coefficient fitting, or development-based selector choice. Finite-pass Flip has nine possible values and can have sparse realized support; adding Drop supplies resolution without guaranteeing useful rankings. The score is an operational heuristic, not a calibrated or monotone measure of perturbation severity.

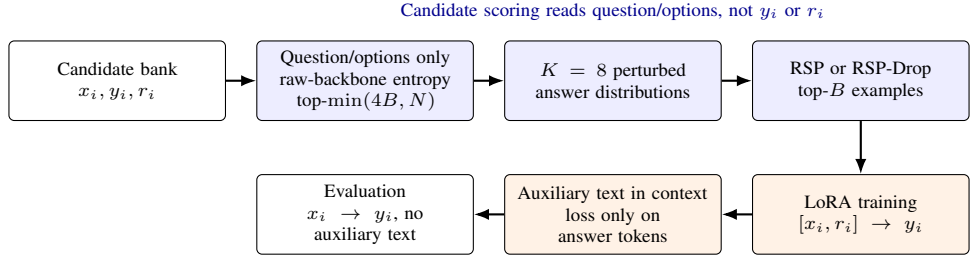
\begin{figure*}[!t]
\centering
\begin{tikzpicture}[
    font=\scriptsize,
    box/.style={draw, rounded corners=2pt, align=center, text width=2.65cm, minimum height=1.05cm, inner sep=3pt},
    blind/.style={box, fill=blue!7},
    train/.style={box, fill=orange!10},
    arrow/.style={-{Latex[length=2mm]}, thick}
]
\node[box] (bank) {Candidate bank\\\(x_i,y_i,r_i\)};
\node[blind, right=4mm of bank] (entropy) {Question/options only\\raw-backbone entropy top-\(\min(4B,N)\)};
\node[blind, right=4mm of entropy] (noise) {\(K=8\) perturbed\\answer distributions};
\node[blind, right=4mm of noise] (score) {RSP or RSP-Drop\\top-\(B\) examples};
\node[train, below=7mm of score] (lora) {LoRA training\\\([x_i,r_i]\rightarrow y_i\)};
\node[train, left=4mm of lora] (loss) {Auxiliary text in context\\loss only on answer tokens};
\node[box, left=4mm of loss] (eval) {Evaluation\\\(x_i\rightarrow y_i\), no auxiliary text};
\draw[arrow] (bank) -- (entropy);
\draw[arrow] (entropy) -- (noise);
\draw[arrow] (noise) -- (score);
\draw[arrow] (score) -- (lora);
\draw[arrow] (lora) -- (loss);
\draw[arrow] (loss) -- (eval);
\node[font=\scriptsize, text=blue!55!black, align=center, text width=7cm, above=1.5mm of noise] {Candidate scoring reads question/options, not \(y_i\) or \(r_i\)};
\end{tikzpicture}
\caption{Shared protocol for \method{} and \droponly{}. Both selectors are label- and auxiliary-text-blind; auxiliary text enters only after selection, as LoRA training context.}
\label{fig:pipeline}
\end{figure*}
\section{Experimental Setup}
\label{sec:experiments}

\subsection{Datasets and Auxiliary-Text Provenance}
\label{sec:data-partitions}

Table~\ref{tab:datasets} lists the fixed study partitions. All normalized records have nonempty questions, options, labels, and auxiliary text, and their UID sets are disjoint. UID disjointness does not establish independence from backbone training or adherence to an original benchmark split.

\begin{table}[!t]
\centering
\caption{Datasets and fixed study partitions. The MedExpQA test count excludes four pool-overlap questions.}
\label{tab:datasets}
\footnotesize
\setlength{\tabcolsep}{3.7pt}
\begin{tabular}{@{}llrrr@{}}
\toprule
Dataset & Answer format & Pool & Dev & Test \\
\midrule
AfriMed-QA & 5 options & 1,500 & 559 & 892 \\
MedExpQA & 4/5 options & 434 & 63 & 121 \\
MedExQA & 4 options & 600 & 164 & 200 \\
PubMedQA & Yes/no/maybe & 600 & 200 & 200 \\
MedMCQA & 4 options & 512 & 256 & 4,096 \\
\bottomrule
\end{tabular}
\end{table}

\paragraph{Auxiliary fields}
AfriMed-QA uses its crowdsourced \texttt{answer\_rationale} \citep{nimo2025afrimedqa}; English MedExpQA uses the doctor-authored, answer-reference-masked \texttt{full\_answer\_no\_ref} \citep{alonso2024medexpqa}. MedExQA uses the first human-validated explanation in its headerless TSV (column 6); column 7 is retained but never used for training \citep{kim2024medexqa}. PubMedQA includes the abstract context in the question and uses \texttt{long\_answer}, the abstract conclusion rather than CoT \citep{jin2019pubmedqa}. MedMCQA uses its expert-explanation field \texttt{exp} \citep{pal2022medmcqa}.

\paragraph{Partition construction}
AfriMed-QA, MedExQA, and PubMedQA deduplicate lowercased, whitespace-normalized questions before their internal splits. AfriMed-QA filters to single-answer, five-option English items, shuffles official training data into pool/development with seed 20260826, and retains official test data. MedExpQA directly maps its official English train/validation/test splits, followed by the test filter below. MedExQA merges 25 development and 940 test items, removes one duplicate, and reshuffles to 600/164/200; its study test is not an untouched official test. PubMedQA reshuffles all 1,000 PQA-L records to 600/200/200 instead of the publication's 500/500 protocol. MedMCQA filters official training records with nonempty \texttt{exp}, excludes 512 earlier-pilot UIDs, and samples 512/256/4,096. It has no explicit question-deduplication step, but no exact normalized-question duplicate or cross-split intersection was recovered in the retained study records.

\paragraph{Overlap and evaluation filtering}
The overlap key lowercases the question, strips its ends, and collapses whitespace. MedExpQA has two pool--development keys, four pool--test keys (five pool rows and four test rows), and no development--test keys. Removing the four overlapping test rows reduces its test size from 125 to 121; one within-test duplicate remains. The pool contains nine duplicated keys covering 19 rows. Across the 21 audited selector--seed selections, each contains one to four pool items overlapping the original test. This post-hoc test filter prevents those duplicated questions from contributing to reported metrics but does not rebuild the partitions or retrain the models. The other four datasets have no recovered exact normalized-question cross-split intersection. Prediction alignment checks are described below.

The \dataset{MedGemma} technical report documents post-training on the training splits of MedMCQA, PubMedQA, MedExpQA, and AfriMed-QA \citep{sellergren2025medgemma}. Thus four benchmark families overlap its documented sources. In particular, the study partitions for PubMedQA and MedMCQA originate from source training data; AfriMed-QA retains its official test, and MedExpQA its official development/test subject to the filter above. MedExQA is not listed among those documented post-training sources, but other contamination is not excluded. Without the unreleased item manifest, exact example overlap cannot be established. These experiments study continued adaptation from a medically post-trained model, not learning in wholly unseen domains; the model card recommends application-specific validation \citep{google2025medgemmacard}.

\subsection{Compared Selectors}

We distinguish information access at selection time. \emph{Question/options-only selectors} comprise Entropy, Entropy-prefiltered Random (\randomfour{}), Margin, BALD-inspired, BatchBALD-inspired, Noise-Stability, SIFT-inspired, NAIT-inspired, \method{}, and \droponly{}. \randomfour{} is the pool-matched control in both the primary comparison (Table~\ref{tab:f1}) and the budget analysis (Table~\ref{tab:rsp-budget-performance}). It samples uniformly from the same top-\(\min(4B,N)\) entropy pool used by both proposed methods, whereas Entropy deterministically retains the most entropic items. The random draw does not inspect explanations and is not unrestricted full-bank random when \(N>4B\).

BALD-inspired uses final-block-input hidden dropout rather than parameter dropout; BatchBALD-inspired uses a sampled joint-disagreement criterion from these stochastic predictions \citep{houlsby2011bayesian,kirsch2019batchbald}. Noise-Stability constructs sensitivity vectors followed by cosine k-center selection. SIFT-inspired combines uncertainty with an embedding-diversity ranking \citep{hubotter2025sift}. NAIT-inspired uses first-to-last prompt-activation deltas across decoder layers, oriented with unlabeled development questions \citep{chen2026nait}. The nominal 5,000-example caps in the implemented diversity comparators do not restrict these pools (\(N\le1500\)); they should not be confused with the \(4B\) entropy pool.

\emph{RSR-inspired, ASLEC-DROP-inspired, and TEMP-inspired} are auxiliary-text-aware comparators \citep{yang2026rsr,wang2026aslec,jin2026temp}: they score candidate explanation text, unlike our two selectors. They are not information-matched controls for question-only scoring. Short names in Table~\ref{tab:f1} denote study-specific implementations, not exact reproductions of the original papers. All use the same retained-set budget and downstream training recipe, which does not imply equal acquisition cost or information access.

BALD-inspired uses eight inverted-Bernoulli hidden-dropout passes with probability 0.1 at the final-block input. Noise-Stability uses eight perturbations with relative RMS scale 0.03, normalized sensitivity vectors, and seeded cosine k-center selection. NAIT-inspired orients one principal component per layer using unlabeled development questions and ranks summed projections. These are operational adaptations of selection criteria.

The retained acquisition protocols additionally record token-rank clipping at 100 for RSR-inspired, deterministic newline/sentence step segmentation for ASLEC-DROP-inspired, and a cosine-diversity weight of 0.5 for SIFT-inspired. TEMP-inspired uses an auxiliary LoRA direction fitted from development explanations. Thus, candidate information access and auxiliary development access are distinct: describing a candidate scorer does not establish label-blindness of every auxiliary stage.

\subsection{Training and Evaluation}
\label{sec:training-evaluation}

Each branch starts from \texttt{google/medgemma-4b-it} with no answer-only warm-up. The code did not pin a Hugging Face revision, so retained artifacts establish the repository ID but not the exact resolved model commit. We initialize a fresh rank-16 LoRA adapter with scale 32 and dropout 0.05 and no bias over \texttt{q/k/v/o/gate/up/down\_proj} modules \citep{hu2022lora}. Training uses two epochs, learning rate \(10^{-4}\), cosine decay, 3\% warm-up, fused AdamW, per-device batch size 1, gradient accumulation 8, gradient checkpointing, and maximum length 1,024. No quantization is requested; loading uses bfloat16 when supported and otherwise float16, but the resolved precision and software package versions were not retained.

\paragraph{Prompts and supervision}
\label{sec:input-handling}
The scoring/evaluation prompt is plain text, without a chat template, and is tokenized with \texttt{add\_special\_tokens=False}:
\begin{quote}\small
\texttt{Please answer the following multiple-choice medical question:\\{}
[QUESTION]\\
A. [OPTION] ...\\
Choose one option. Return only the final option letter.\\
Final Answer:}
\end{quote}
The training serialization is
\begin{quote}\small
\texttt{[question, options] \textbackslash nReasoning:\textbackslash n [auxiliary text]\\
Using the reasoning above, return only the final option letter.\\
Final Answer: [LETTER]}
\end{quote}
followed by EOS. Only the leading-space answer letter, newline, and EOS target tokens receive loss. Evaluation provides the question and options followed by the answer instruction, then chooses among legal next-letter logits; there is no free generation and therefore no invalid-output category. Accuracy and Macro-F1 use answer-option positions (A--C, A--D, or A--E) as classes, not medical diagnosis classes. We report mean and sample standard deviation across branch seeds 13, 23, and 37. For stochastic selectors, including \method{} and \randomfour{}, a branch seed determines both acquisition randomness and downstream training; fixed selections vary only in training. All methods share each dataset's fixed pool, development, and test partitions. Slurm manifests request GPUs labeled RTX5000 for \method{}, BALD-inspired, Noise-Stability, and \randomfour{}, and RTX6000 for NAIT-inspired. They record requested GRES classes rather than per-job GPU UUIDs, so the experiment families are not hardware-uniform or bitwise reproducible.

\paragraph{Cleaning and truncation}
Three regular expressions remove newline-delimited tails of the forms \texttt{final answer}, \texttt{therefore ... answer is}, and \texttt{correct option}; other answer cues are not masked. The target \texttt{" [LETTER]\textbackslash n"} and EOS are fully reserved. When length exceeds the limit, the question/options prefix is truncated from its middle, preserving its head and tail, followed by right-truncation of auxiliary text; the instruction and full target remain. Evaluation uses head--tail truncation to 1,024 tokens. No question or auxiliary-text truncation was needed in the 15 primary \(B=128\) RSP branches; this check does not cover all comparator branches.

\paragraph{Prediction alignment and aggregation}
\label{sec:prediction-alignment}
The primary per-seed metrics are recomputed from retained per-question predictions using the common test-label order and the MedExpQA filter. Original NPZ files contain \texttt{test\_probs} and \texttt{test\_gold} but not test UIDs, so positional pairing relies on the frozen split order. Before filtering, checks compare the test length, legal-answer-space width, and complete ordered gold vector against the hashed source test manifest; gold-vector agreement is not independent evidence of UID alignment. Dataset averages use equal weights. The paired summaries in RQ4 are recomputed from the retained unrounded per-dataset/seed metrics for the same result families, with MedExpQA already filtered to 121 items. No second filtering is applied; sample SD describes the three paired branch differences and is not a confidence interval.

\subsection{Computational Resource Inventory}
\label{sec:cost-inventory}
Automatic \(\lambda_0\) requires no adapter fit or development evaluation. Scoring uses one clean and eight perturbed passes per candidate; every downstream branch trains for two epochs. The RSP-Drop budget evaluation comprises 45 fits using retained score components. Table~\ref{tab:computational-cost} covers only the 90 completed RSP/Random@\(4B\) budget branches, not RSP-Drop or the other comparators.

Training time is recorded \texttt{Trainer.train\_runtime}, excluding loading, final saving, evaluation, and queueing. Non-padding input-token exposure is reconstructed on CPU from each branch's retained tokenizer, serializer, selected order, cleaning and truncation, over two epochs; historical tokenizer identity was not independently pinned for reused \(B=128\) runs. Progress logs confirm 16/64 optimizer updates at \(B=64/256\); \(B=128\) progress records are unavailable. Entropy-cache construction, extra perturbation scoring, and sorting/sampling times cannot be separated, so scoring time is unavailable, not zero. These are descriptive records, not controlled timing or end-to-end efficiency comparisons.

\begin{table}[!t]
\centering
\caption{Resource inventory: mean per run across five datasets and three seeds (15 runs per row). Training time is measured; optimizer updates are measured where available; input tokens are reconstructed.}
\label{tab:computational-cost}
\scriptsize
\setlength{\tabcolsep}{3pt}
\begin{tabular}{@{}clrrr@{}}
\toprule
\(B\) & Method & \shortstack{Training\\seconds} & \shortstack{Input tokens\\\((10^3)\)} & \shortstack{Optimizer\\updates} \\
\midrule
64 & \method{} & 43.55 & 32.47 & 16 \\
64 & \randomfour{} & 43.48 & 34.45 & 16 \\
128 & \method{} & 86.72 & 64.42 & NA \\
128 & \randomfour{} & 87.30 & 69.41 & NA \\
256 & \method{} & 172.86 & 130.96 & 64 \\
256 & \randomfour{} & 172.78 & 137.58 & 64 \\
\bottomrule
\end{tabular}
\par\smallskip
\parbox{\columnwidth}{\scriptsize Scoring time is unavailable for all rows. Tokens include the full non-padding input and target over both epochs. Both selectors require entropy scoring; cache reuse is not zero cost. These jobs requested one RTX5000-class GPU; actual devices and precision were not retained.}
\end{table}

\section{Results}
\label{sec:results}

Table~\ref{tab:f1} reports the mixed-overlap benchmark comparison with both proposed selectors and the broader baseline set. Each dataset cell reports mean Accuracy / Macro-F1; the final column averages the five dataset means with equal weight. All selected sets are restored to original pool-index order before training. The fixed item budget does not equalize explanation length, selector information access, or acquisition cost.

\begin{table*}[!t]
\centering
\caption{Overall performance at \(B=128\): test Accuracy / Macro-F1 (\%, three-seed means). Dataset-macro avg.\ is the unweighted mean across the five datasets, computed before rounding. Bold marks the largest mean separately for each metric and column, not statistical significance. Seed-level dataset averages are in Table~\ref{tab:primary-seed-summary}.}
\label{tab:f1}
\footnotesize
\setlength{\tabcolsep}{3pt}
\begin{tabular}{lcccccc}
\toprule
Method & AfriMed-QA & MedExpQA & MedExQA & PubMedQA & MedMCQA & \shortstack{Dataset-macro\\avg.} \\
\midrule
\multicolumn{7}{l}{\emph{Question/options-only controls}} \\
\randomfour{} & 49.36 / 46.04 & 51.79 / 49.64 & 50.33 / 49.64 & 69.50 / 48.58 & 36.09 / 31.68 & 51.42 / 45.11 \\
Entropy & 50.34 / 47.19 & 51.24 / 49.00 & 48.67 / 48.60 & 70.00 / 50.74 & 38.00 / 33.52 & 51.65 / 45.81 \\
Margin & 51.49 / 49.34 & 52.07 / 48.63 & 48.67 / 48.54 & 69.00 / 49.17 & 38.48 / 34.72 & 51.94 / 46.08 \\
BALD & 53.33 / 50.86 & 50.96 / 46.99 & 48.67 / 47.83 & 69.17 / 48.79 & 37.64 / 35.64 & 51.95 / 46.02 \\
BatchBALD & 49.96 / 46.66 & 50.14 / 45.88 & 49.50 / 49.19 & \textbf{71.67} / \textbf{54.34} & 37.65 / 35.00 & 51.78 / 46.21 \\
Noise-Stability & 48.51 / 43.50 & 50.96 / 49.02 & 45.33 / 45.62 & 70.00 / 48.94 & 39.25 / 37.49 & 50.81 / 44.92 \\
SIFT & 53.44 / \textbf{51.29} & 50.41 / 47.39 & 47.33 / 47.13 & 68.50 / 48.95 & 37.73 / 34.00 & 51.48 / 45.75 \\
NAIT & 50.22 / 45.95 & 52.89 / 48.54 & 41.83 / 41.57 & 54.67 / 25.80 & 36.69 / 32.57 & 47.26 / 38.89 \\
\midrule
\multicolumn{7}{l}{\emph{Auxiliary-text-aware comparators}} \\
RSR & 50.67 / 47.23 & 49.59 / 46.74 & 51.33 / 51.31 & 69.67 / 49.70 & \textbf{40.60} / \textbf{38.80} & 52.37 / 46.76 \\
ASLEC-DROP & 51.53 / 48.62 & 52.62 / 50.64 & 50.17 / 50.10 & 71.17 / 50.07 & 38.66 / 36.58 & 52.83 / 47.20 \\
TEMP & 50.78 / 47.08 & 52.89 / 49.73 & 51.17 / 49.75 & 70.00 / 49.23 & 37.77 / 34.27 & 52.52 / 46.01 \\
\midrule
\multicolumn{7}{l}{\emph{Perturbation-based selection rules (question/options-only)}} \\
\method{} & \textbf{53.70} / 51.26 & 51.24 / 49.48 & 52.50 / 52.31 & 70.00 / 51.00 & 37.54 / 33.72 & 53.00 / 47.55 \\
\droponly{} & 52.39 / 50.11 & \textbf{53.72} / \textbf{50.82} & \textbf{54.50} / \textbf{54.28} & 68.83 / 49.92 & 38.56 / 36.28 & \textbf{53.60} / \textbf{48.28} \\
\bottomrule
\end{tabular}
\par\smallskip
\parbox{0.98\textwidth}{\scriptsize BALD, BatchBALD, SIFT, NAIT, RSR, ASLEC-DROP, and TEMP denote the study-specific inspired implementations. RSR, ASLEC-DROP, and TEMP read auxiliary text during selection; all other listed selectors are question/options-only. MedExpQA excludes four pool-overlap test questions (\(n=121\)). Classes are answer-option positions.}
\end{table*}

\subsection{RQ1: Overall Performance at the Primary Budget}

At \(B=128\), \droponly{} obtains the largest dataset-averaged Accuracy/Macro-F1 among the reported selectors (53.60/48.28), followed by \method{} (53.00/47.55). These equal-weight averages summarize the five evaluated benchmarks, rather than a pooled patient-level metric.

Dataset-level results are heterogeneous. \droponly{} has the largest mean Accuracy and Macro-F1 on MedExpQA (53.72/50.82) and MedExQA (54.50/54.28). On AfriMed-QA, \method{} has the highest accuracy (53.70), whereas SIFT-inspired has a slightly higher Macro-F1 (51.29 versus 51.26). BatchBALD-inspired leads on PubMedQA (71.67/54.34), and RSR-inspired leads on MedMCQA (40.60/38.80).

In the direct comparison, \droponly{} is higher on both metrics on MedExpQA, MedExQA, and MedMCQA, whereas \method{} is higher on AfriMed-QA and PubMedQA. On MedMCQA, \droponly{} improves on \method{} by 1.02 pp accuracy and 2.56 pp Macro-F1, but remains below Noise-Stability, ASLEC-DROP-inspired, and RSR-inspired. The averages therefore support the utility of the two selection rules under this protocol, not uniform superiority across datasets.

\paragraph{Seed-level variation}
Table~\ref{tab:primary-seed-summary} retains each branch's five-dataset average for the primary comparison; it complements the per-dataset means in Table~\ref{tab:f1}. Each seed average is equally weighted across datasets before rounding.

\begin{table}[!t]
\centering
\caption{Primary-budget dataset averages by branch seed: Accuracy / Macro-F1 (\%), \(B=128\). These are seed-level summaries, not additional runs.}
\label{tab:primary-seed-summary}
\scriptsize
\setlength{\tabcolsep}{2.1pt}
\begin{tabular}{@{}lccc@{}}
\toprule
Method & Seed 13 & Seed 23 & Seed 37 \\
\midrule
\randomfour{} & 51.60 / 45.56 & 50.23 / 43.89 & 52.42 / 45.90 \\
Entropy & 51.24 / 45.46 & 53.29 / 47.93 & 50.41 / 44.04 \\
Margin & 52.35 / 46.78 & 50.65 / 45.16 & 52.83 / 46.31 \\
BALD & 52.36 / 45.55 & 51.36 / 45.62 & 52.13 / 46.90 \\
BatchBALD & 53.90 / 50.06 & 50.86 / 43.79 & 50.59 / 44.79 \\
Noise-Stability & 49.48 / 44.16 & 51.32 / 45.17 & 51.63 / 45.41 \\
SIFT & 51.63 / 46.62 & 52.08 / 45.31 & 50.73 / 45.32 \\
NAIT & 47.97 / 39.95 & 47.89 / 40.27 & 45.92 / 36.44 \\
RSR & 52.37 / 46.82 & 52.96 / 47.14 & 51.79 / 46.31 \\
ASLEC-DROP & 53.24 / 47.47 & 52.60 / 46.73 & 52.64 / 47.39 \\
TEMP & 51.85 / 45.55 & 52.51 / 46.01 & 53.21 / 46.49 \\
\method{} & 52.75 / 47.20 & 52.93 / 47.39 & 53.31 / 48.08 \\
\droponly{} & 53.65 / 49.01 & 53.37 / 47.83 & 53.79 / 48.00 \\
\bottomrule
\end{tabular}
\end{table}

\paragraph{No-update reference}
\label{sec:raw-reference}
A single raw-checkpoint inference run gives Macro-F1 38.13, 37.12, 35.08, 31.75, and 29.14 on AfriMed-QA, MedExpQA, MedExQA, PubMedQA, and MedMCQA, respectively. These values have no training-seed SD and do not constitute a matched answer-only adaptation control.

\subsection{RQ2: Comparison with Pool-Matched Random Selection}

The pool-matched control in Table~\ref{tab:f1} tests the value of sensitivity reranking within the same entropy-restricted candidate pool. At \(B=128\), dataset-averaged Macro-F1 is 45.11 for \randomfour{}, 47.55 for \method{}, and 48.28 for \droponly{}, corresponding to descriptive improvements of 2.44 and 3.17 percentage points.

\droponly{} exceeds \randomfour{} on both metrics on AfriMed-QA, MedExpQA, MedExQA, and MedMCQA. On PubMedQA, its accuracy is lower (68.83 versus 69.50), but Macro-F1 is higher (49.92 versus 48.58). \method{} exceeds the random control on both metrics on four datasets, with lower means on MedExpQA. Table~\ref{tab:paired-budget} adds paired branch-level differences rather than inferring uncertainty from separate method SDs.

At this budget, MedExpQA and MedMCQA have \(N\le4B\), so the matched control is also full-pool uniform random. For the other three datasets, the primary comparison does not evaluate unrestricted random selection. The prefilter itself is examined separately in RQ5.

\subsection{RQ3: Component Utility and Selection Behavior}

Both reported methods use the current budget-evaluation result family; earlier RSP-Drop runs are not mixed into Table~\ref{tab:f1}. Their comparison changes the inclusion of Flip in the ranking rule while retaining the same entropy-pool definition and adaptation recipe.

Across the 45 audited candidate-pool records (five datasets, \(B=64,128,256\), and seeds 13/23/37), \(F_i\) takes only the values 0 and 1, although nine values are possible at \(K=8\). In the 15 audited \(B=128\) selected-set comparisons across all five datasets and three seeds, the RSP-exclusive samples have \(F_i=1,D_i=0\), whereas the RSP-Drop-exclusive samples have \(F_i=0,D_i>0\). Adding Flip therefore changes membership by introducing a distinct group of candidates, rather than merely refining their contraction ranking.

The MedMCQA \(B=128\), seed-13 case study quantifies this exchange (Table~\ref{tab:selection-case}). Define \(k=|\mathcal S_{\mathrm{RSP}}\setminus\mathcal S_{\mathrm{Drop}}|\). Because both sets contain 128 examples, each side has \(k=26\) exclusive examples, the intersection contains 102 examples (79.69\% of either set), and the symmetric difference contains 52 examples. The clean-margin means of the two exclusive groups are 0.098 and 0.707. Thus, RSP exchanges some contraction-selected examples for examples closer to the clean prediction boundary in this audited case. A qualitatively similar margin contrast is recorded for AfriMed-QA at \(B=128\), seed 13, where RSP instead has higher three-seed \(B=128\) performance. The contrast describes which examples are selected; it does not explain the downstream difference or provide a rule for choosing between selectors.

\begin{table}[!t]
\centering
\caption{Selected-set comparison on MedMCQA, \(B=128\), seed 13. Statistics concern each selector's exclusive examples, not its complete selected set.}
\label{tab:selection-case}
\scriptsize
\setlength{\tabcolsep}{4pt}
\begin{tabular}{lrrrc}
\toprule
Exclusive group & \(n\) & \(F_i\) & \(D_i\) & \shortstack{Mean clean\\margin} \\
\midrule
\method{} only & 26 & 1 & 0 & 0.098 \\
\droponly{} only & 26 & 0 & \(>0\) & 0.707 \\
\bottomrule
\end{tabular}
\end{table}

No selection-cutoff ties occur for either rule at \(B=64,128\). At \(B=256\), both rules have zero-score cutoff ties on all datasets except AfriMed-QA (12 dataset--budget--seed settings). Removing Flip therefore does not generally eliminate tie-driven selection. RMS balancing matches empirical component scales; it does not calibrate their downstream utility.

\subsection{RQ4: Sensitivity to the Retained-Set Budget}
\label{sec:budget}

Table~\ref{tab:rsp-budget-performance} compares \method{}, \droponly{}, and their pool-matched random control at \(B=64,128,256\). Both sensitivity rules have higher dataset-averaged Accuracy and Macro-F1 than \randomfour{} at each budget. Between the two rules, \droponly{} has higher mean Macro-F1 at all three budgets (42.03, 48.28, and 50.43 versus 41.41, 47.55, and 50.37), although the \(B=256\) means are especially close. This describes the evaluated settings, not a statistically established or universal ordering.

To show branch variability, for methods \(A\) and \(C\) and metric \(q\), we compute
\begin{equation}
\delta_{B,s}^{A-C}=\frac{1}{5}\sum_d
\left(q_{A,d,B,s}-q_{C,d,B,s}\right).
\label{eq:paired-budget-delta}
\end{equation}
Table~\ref{tab:paired-budget} reports the mean and sample SD of the three paired values, computed from unrounded retained metrics before display rounding. At \(B=128\), the RSP-Drop-minus-RSP Macro-F1 differences are \(+1.82,+0.44,-0.08\) pp: a higher mean does not imply a gain on every seed. At \(B=256\), the corresponding mean is \(+0.06\) pp with SD 1.67 pp, so the near-tie in averages should not be interpreted as established superiority or equivalence.

Dataset-level directions also differ. Both metrics favor RSP-Drop over RSP on MedMCQA at all three budgets and RSP over RSP-Drop on PubMedQA at all three. MedExpQA favors RSP-Drop at 64/128 but RSP at 256; MedExQA favors RSP at 64 but RSP-Drop at 128/256. AfriMed-QA favors RSP at 128, while the other budgets show metric-dependent directions or small gaps. At \(B=256\), \randomfour{} has the highest AfriMed-QA Accuracy and Macro-F1 and the highest MedExpQA Accuracy among the three methods.

At \(B=64\), every entropy pool contains 256 candidates. At \(B=128\), MedExpQA and MedMCQA already include their full banks. At \(B=256\), all banks except AfriMed-QA are fully included; AfriMed-QA retains 1,024 of 1,500 candidates. Fixed two-epoch training increases optimizer updates and token exposure with \(B\). Accordingly, the within-budget comparisons hold retained-set size and recipe fixed, whereas cross-budget changes do not isolate selection effects.

\begin{table}[!t]
\centering
\caption{Paired dataset-macro differences from unrounded retained metrics (percentage points; mean \(\pm\) sample SD across seeds 13/23/37). Each comparison is first method minus second.}
\label{tab:paired-budget}
\scriptsize
\setlength{\tabcolsep}{2.5pt}
\begin{tabular}{clrr}
\toprule
\(B\) & Comparison & Accuracy & Macro-F1 \\
\midrule
64 & \method{} -- \randomfour{} & $+1.12\!\pm\!1.22$ & $+0.60\!\pm\!1.52$ \\
64 & \droponly{} -- \randomfour{} & $+1.02\!\pm\!1.18$ & $+1.22\!\pm\!0.89$ \\
64 & \droponly{} -- \method{} & $-0.10\!\pm\!0.71$ & $+0.62\!\pm\!0.69$ \\
\midrule
128 & \method{} -- \randomfour{} & $+1.58\!\pm\!0.99$ & $+2.44\!\pm\!0.96$ \\
128 & \droponly{} -- \randomfour{} & $+2.18\!\pm\!0.90$ & $+3.17\!\pm\!0.95$ \\
128 & \droponly{} -- \method{} & $+0.60\!\pm\!0.25$ & $+0.73\!\pm\!0.98$ \\
\midrule
256 & \method{} -- \randomfour{} & $+0.21\!\pm\!0.79$ & $+0.87\!\pm\!1.25$ \\
256 & \droponly{} -- \randomfour{} & $+0.84\!\pm\!1.87$ & $+0.93\!\pm\!1.76$ \\
256 & \droponly{} -- \method{} & $+0.63\!\pm\!1.50$ & $+0.06\!\pm\!1.67$ \\
\bottomrule
\end{tabular}
\par\smallskip
\parbox{\columnwidth}{\scriptsize Differences are computed within each dataset and seed, then averaged equally over five datasets before calculating the three-seed sample SD. SD is not a confidence interval. Prediction pairing retains the frozen-order assumption described in Section~\ref{sec:training-evaluation}.}
\end{table}

\begin{table*}[!t]
\centering
\caption{Budget sensitivity of RSP, RSP-Drop, and entropy-prefiltered Random: Accuracy / Macro-F1 (\%, three-seed means). All three use the same top-\(\min(4B,N)\) entropy pool within each budget. Bold marks the largest mean among the three methods for each metric, dataset, and budget. Dataset-macro avg.\ uses equal dataset weight. Paired branch variability is summarized in Table~\ref{tab:paired-budget}.}
\label{tab:rsp-budget-performance}
\footnotesize
\setlength{\tabcolsep}{3pt}
\begin{tabular}{clcccccc}
\toprule
\(B\) & Method & AfriMed-QA & MedExpQA & MedExQA & PubMedQA & MedMCQA & \shortstack{Dataset-macro\\avg.} \\
\midrule
64 & \method{} & \textbf{48.43} / 43.52 & 45.73 / 41.35 & \textbf{44.50} / \textbf{44.37} & \textbf{68.17} / \textbf{46.75} & 35.42 / 31.06 & \textbf{48.45} / 41.41 \\
64 & \droponly{} & 47.83 / \textbf{44.13} & \textbf{48.21} / \textbf{45.29} & 44.17 / 44.16 & 65.17 / 44.39 & \textbf{36.39} / \textbf{32.19} & 48.35 / \textbf{42.03} \\
64 & \randomfour{} & 48.13 / 44.12 & 45.45 / 42.34 & 43.00 / 42.70 & 64.00 / 43.18 & 36.07 / 31.73 & 47.33 / 40.81 \\
\midrule
128 & \method{} & \textbf{53.70} / \textbf{51.26} & 51.24 / 49.48 & 52.50 / 52.31 & \textbf{70.00} / \textbf{51.00} & 37.54 / 33.72 & 53.00 / 47.55 \\
128 & \droponly{} & 52.39 / 50.11 & \textbf{53.72} / \textbf{50.82} & \textbf{54.50} / \textbf{54.28} & 68.83 / 49.92 & \textbf{38.56} / \textbf{36.28} & \textbf{53.60} / \textbf{48.28} \\
128 & \randomfour{} & 49.36 / 46.04 & 51.79 / 49.64 & 50.33 / 49.64 & 69.50 / 48.58 & 36.09 / 31.68 & 51.42 / 45.11 \\
\midrule
256 & \method{} & 53.03 / 51.28 & 53.44 / \textbf{51.80} & 57.00 / 56.35 & \textbf{69.83} / \textbf{51.06} & 42.91 / 41.36 & 55.24 / 50.37 \\
256 & \droponly{} & 53.10 / 51.20 & 52.34 / 49.22 & \textbf{60.17} / \textbf{59.52} & 69.17 / 48.77 & \textbf{44.61} / \textbf{43.42} & \textbf{55.88} / \textbf{50.43} \\
256 & \randomfour{} & \textbf{54.07} / \textbf{52.34} & \textbf{53.99} / 51.00 & 54.67 / 53.54 & 69.33 / 48.88 & 43.10 / 41.72 & 55.03 / 49.50 \\
\bottomrule
\end{tabular}
\end{table*}

\subsection{RQ5: Effect of Entropy Prefiltering}
\label{sec:entropy-ablation}

We compare entropy-prefiltered \method{} and \droponly{} with the reported matched full-pool counterparts under the current perturbation and training protocol. Table~\ref{tab:entropy-ablation} includes the nine dataset--budget settings where top-\(4B\) removes candidates. Settings with \(\min(4B,N)=N\) are excluded: the selectors then produce the same selected UID order, so differences between separate training runs cannot identify a prefiltering effect.

The frozen RSP-Drop preparation ranks the complete bank directly by \(D_i\) for the full-pool condition. Its shared candidates reuse the retained per-item components; missing candidates were scored in the recorded budget supplements, with the remaining AfriMed-QA bank completed by continuing the inherited Gaussian stream. This describes the historical score construction, not a new scoring run. The resulting comparison concerns the complete implemented selection procedures.

Full-pool RSP reuses this completed bank of \(F_i,D_i\) components and recomputes \(a_F\), \(a_D\), and \(\lambda_0\) over all candidates separately for each dataset and seed, rather than retaining the entropy-prefiltered coefficient. The resulting coefficient is shared across budgets within that dataset--seed pair. Final scores are computed over the complete bank, and the top \(B\) candidates are selected with ties resolved by original pool index.

For each method, prefiltering increases Accuracy in five of the nine active settings and Macro-F1 in six. At \(B=128\), mean Macro-F1 increases for both methods on all three active datasets. Accuracy is not uniformly non-negative: on PubMedQA, RSP-Drop decreases by 1.17 pp despite a 0.62-point Macro-F1 increase. At \(B=256\), the only active comparison is AfriMed-QA, where both methods favor the full pool on both metrics. Effects at \(B=64\) are mixed.

RSP's largest Macro-F1 increase is on AfriMed-QA at \(B=128\) (+2.62 pp), whereas its largest Accuracy increase is on MedExQA at \(B=64\) (+2.17 pp); its largest Macro-F1 decrease is on MedMCQA at \(B=64\) (-2.49 pp). For RSP-Drop, the largest increases occur on MedExQA at \(B=128\) (+2.83 Accuracy and +3.41 Macro-F1 pp), and the largest decreases on PubMedQA at \(B=64\) (-2.50 and -2.19 pp). Thus, entropy restriction is neither uniformly helpful nor uniformly harmful. The results compare the implemented selection procedures with and without the prefilter; they do not establish statistical significance. The full-pool RSP ablation comprises 27 training branches across these nine active settings; its coefficients are positive (2.982--3.728), so the zero-coefficient branch is not exercised.

\begin{table*}[!t]
\centering
\caption{Entropy-prefilter ablation for RSP and RSP-Drop on settings where top-\(4B\) is smaller than the full pool. Entries are Accuracy / Macro-F1 (\%, three-seed means); \(\Delta\) is entropy-prefiltered minus full-pool performance in percentage points. Differences are computed before rounding and may therefore differ slightly from subtraction of the displayed means. Separate training runs share seeds and the frozen recipe, but the analysis is descriptive rather than a significance test.}
\label{tab:entropy-ablation}
\scriptsize
\setlength{\tabcolsep}{3.1pt}
\begin{tabular}{lllrrrr}
\toprule
\(B\) & Dataset & Method & \shortstack{Top-\(4B\) /\ full pool} & Entropy-prefiltered & Full-pool & \(\Delta\) \\
\midrule
64 & AfriMed-QA & RSP & 256 / 1,500 & 48.43 / 43.52 & 46.82 / 42.09 & +1.61 / +1.44 \\
 & & RSP-Drop & & 47.83 / 44.13 & 46.79 / 42.53 & +1.05 / +1.59 \\
64 & MedExpQA & RSP & 256 / 434 & 45.73 / 41.35 & 46.01 / 42.33 & -0.28 / -0.98 \\
 & & RSP-Drop & & 48.21 / 45.29 & 46.28 / 42.32 & +1.93 / +2.97 \\
64 & MedExQA & RSP & 256 / 600 & 44.50 / 44.37 & 42.33 / 42.61 & +2.17 / +1.76 \\
 & & RSP-Drop & & 44.17 / 44.16 & 44.33 / 43.51 & -0.17 / +0.65 \\
64 & PubMedQA & RSP & 256 / 600 & 68.17 / 46.75 & 66.50 / 44.71 & +1.67 / +2.04 \\
 & & RSP-Drop & & 65.17 / 44.39 & 67.67 / 46.59 & -2.50 / -2.19 \\
64 & MedMCQA & RSP & 256 / 512 & 35.42 / 31.06 & 36.98 / 33.55 & -1.56 / -2.49 \\
 & & RSP-Drop & & 36.39 / 32.19 & 34.70 / 32.23 & +1.69 / -0.04 \\
\midrule
128 & AfriMed-QA & RSP & 512 / 1,500 & 53.70 / 51.26 & 51.72 / 48.64 & +1.98 / +2.62 \\
 & & RSP-Drop & & 52.39 / 50.11 & 51.05 / 48.24 & +1.35 / +1.87 \\
128 & MedExQA & RSP & 512 / 600 & 52.50 / 52.31 & 52.17 / 52.18 & +0.33 / +0.13 \\
 & & RSP-Drop & & 54.50 / 54.28 & 51.67 / 50.87 & +2.83 / +3.41 \\
128 & PubMedQA & RSP & 512 / 600 & 70.00 / 51.00 & 70.00 / 50.91 & +0.00 / +0.09 \\
 & & RSP-Drop & & 68.83 / 49.92 & 70.00 / 49.31 & -1.17 / +0.62 \\
\midrule
256 & AfriMed-QA & RSP & 1,024 / 1,500 & 53.03 / 51.28 & 54.93 / 53.15 & -1.91 / -1.86 \\
 & & RSP-Drop & & 53.10 / 51.20 & 54.67 / 52.96 & -1.57 / -1.77 \\
\bottomrule
\end{tabular}
\end{table*}

\section{Limitations and Scope}
\label{sec:limitations}

\paragraph{Causal scope}
All selected examples include auxiliary explanations during training, whereas evaluation omits them. Without a matched answer-only adaptation control, neither the selector comparisons nor the single no-update reference (RQ1) isolates the contribution of explanation conditioning. This train--test context shift is shared by the compared adaptation protocols. The prefilter ablation concerns only RSP and RSP-Drop; unrestricted random selection is not separately evaluated on every restricted-pool setting.

\paragraph{Selection validity}
Automatic \(\lambda_0\) balances component scales without establishing their downstream usefulness. Finite-pass Flip frequencies are coarse, and the fixed absolute \(\sigma\) is not normalized to hidden-state magnitude. We study one backbone, one LoRA recipe, three seeds, and three retained-set budgets. Retrospective budget and component analyses on viewed tests do not establish confirmatory validity. RSP-Drop was investigated after earlier results were viewed; its inclusion as a reported method does not make that comparison prospective. We have not validated a rule for choosing between the two selectors on unseen datasets.

\paragraph{Data and generalization}
Auxiliary provenance is heterogeneous: dataset-provided human explanations, doctor-authored reference text, an article conclusion, and an expert-explanation field are not interchangeable. The narrow preprocessing rules remove some explicit final-answer tails but do not constitute a comprehensive answer-leakage or factuality audit. We do not observe the latent properties that make one selected subset more useful than another, so the component ablation cannot attribute downstream differences uniquely to ambiguity, label correctness, auxiliary-text characteristics, or any other candidate property. MedExpQA has normalized question-text overlap between pool and development/test. We remove pool-overlap questions from its reported test metrics, but this post-hoc filter is not a full pool/dev/test reconstruction or retraining; one within-test duplicate also remains. \dataset{MedGemma} documents post-training on four benchmark training families used here, and exact item overlap is unresolved. The five-dataset analysis therefore mixes backbone-overlap conditions. We have not demonstrated transfer across model revisions, other backbones, budgets beyond those evaluated, languages, institutions, or newly collected data.

\paragraph{Clinical scope and reproducibility}
These public medical-knowledge benchmarks do not measure patient outcomes, harmful clinical errors, calibration, fairness, or deployment readiness. Neither selector assesses explanation faithfulness or factuality. Historical model revisions and software versions were not fully pinned, and GPU classes differ between some branches. Separate retraining runs can exhibit stochastic variation even when selected-example membership is unchanged; identical selections are therefore not treated as independent evidence for a prefiltering effect. Future clinical use would require appropriate governance and independent clinical validation.

\section{Conclusion}

We study perturbation-based data selection for medical QA using two ranking rules: \method{} combines prediction changes with non-flipping margin contraction, while \droponly{} uses contraction alone after the same entropy prefilter. At \(B=128\), their dataset-averaged Macro-F1 scores are 47.55 and 48.28, compared with 45.11 for pool-matched random selection. Both rules retain positive dataset-averaged differences over that control across the three evaluated budgets, while their relative performance varies across datasets and metrics.

The selected-set case study shows that adding Flip exchanges contraction-selected examples for a group with a different clean-margin profile. The full-pool ablation further shows that entropy restriction can help or hurt depending on the setting. Together, these results identify the component choice, candidate-pool restriction, and retained budget as distinct aspects of the selection procedure. They do not establish a universal selector-choice rule. Evaluation on independent data and additional backbones, with fully tracked inference and training configurations, is needed to assess transfer beyond the present retrospective study.

\FloatBarrier
\bibliographystyle{IEEEtranN}
\bibliography{sample}

\end{document}